\documentclass[10pt,twocolumn,letterpaper]{article}

\usepackage{cvpr}
\usepackage{times}
\usepackage{epsfig}
\usepackage{graphicx}
\usepackage{amsmath}
\usepackage{amssymb}
\usepackage{graphicx}
\usepackage{subcaption}
\usepackage{float}
\usepackage{caption}
\usepackage{lscape}                                         % Useful for wide tables or figures.

\usepackage[lined,ruled,linesnumbered]{algorithm2e}

\usepackage{booktabs}                   % Publication quality tables
\usepackage{multirow}

\usepackage{paralist}
\usepackage{enumitem}

\usepackage{bm}                          % Make bold, italic math symbols
\usepackage{epsfig}                      % for figures
\usepackage{graphicx}                  % another package that works for figures
\usepackage{times}
\usepackage{mathptmx}
\usepackage{mathtools}
\usepackage{amssymb,amsmath}   % Short math guide for LaTeX ftp://ftp.ams.org/pub/tex/doc/amsmath/short-math-guide.pdf
\usepackage{scrextend}
\usepackage{tablefootnote}

\usepackage{units}

\usepackage{eucal}

\usepackage{comment}

\usepackage{url}  % Hyphenation of URLs.
\usepackage{xspace}
\usepackage{setspace}

\usepackage{layout}

\usepackage{amsmath,amsfonts,bm}

\def\eqref#1{equation~\ref{#1}}
\def\1{\bm{1}}

\DeclareMathAlphabet{\mathsfit}{\encodingdefault}{\sfdefault}{m}{sl}
\SetMathAlphabet{\mathsfit}{bold}{\encodingdefault}{\sfdefault}{bx}{n}

\newlength\paramargin
\newlength\figmargin
\newlength\secmargin
\newlength\figcapmargin

\long\def\ignorethis#1{}

\newcommand{\comments}{0}

\ifthenelse{\equal{\comments}{1}}
{
\newcommand {\yencheng}[1]{{\color{magenta}\textbf{Yen-Cheng: }#1}\normalfont}
\newcommand {\zk}[1]{{\color{blue}\textbf{Zsolt: }#1}\normalfont}
\newcommand {\jtian}[1]{{\color{blue}\textbf{JTian: }#1}\normalfont}
\newcommand {\cwkuo}[1]{{\color{blue}\textbf{Albert: }#1}\normalfont}
\newcommand {\cyma}[1]{{\color{cyan}{\textbf{Kevin: }#1}\normalfont}}
}
{
\newcommand {\zk}[1]{{\color{blue}{}}\normalfont}
\newcommand {\jtian}[1]{{\color{blue}{}}\normalfont}
\newcommand {\cwkuo}[1]{{\color{blue}{}}\normalfont}
\newcommand {\cyma}[1]{{\color{cyan}{{}}\normalfont}}
}

\newcommand{\final}{0}
\ifthenelse{\equal{\final}{1}}
{
\renewcommand{\yencheng}[1]{}
\renewcommand{\cwkuo}[1]{}
\renewcommand{\jtian}[1]{}
\renewcommand{\zsolt}[1]{}
\renewcommand{\cyma}[1]{}
}
{}

\usepackage[pagebackref=true,breaklinks=true,letterpaper=true,colorlinks,bookmarks=false]{hyperref}

\cvprfinalcopy % *** Uncomment this line for the final submission
\def\cvprPaperID{1359} % *** Enter the CVPR Paper ID here

\ifcvprfinal\fi

\begin{document}

%%%%%%%%% TITLE
\title{When2com: Multi-Agent Perception via Communication Graph Grouping}
\author{Yen-Cheng Liu, Junjiao Tian\thanks{equal contribution}, Nathaniel Glaser\footnotemark[1],  Zsolt Kira\\
Georgia Institute of Technology\\
{\tt\small \{ycliu, jtian73, nglaser3, zkira\}@gatech.edu}
% For a paper whose authors are all at the same institution,
% omit the following lines up until the closing ``}''.
% Additional authors and addresses can be added with ``\and'',
% just like the second author.
% To save space, use either the email address or home page, not both
}

\maketitle

%%%%%%%%% ABSTRACT
\begin{abstract}

While significant advances have been made for single-agent perception, many applications require multiple sensing agents and cross-agent communication due to benefits such as coverage and robustness. 
It is therefore critical to develop frameworks which support multi-agent collaborative perception in a distributed and bandwidth-efficient manner. 
In this paper, we address the collaborative perception problem, where one agent is required to perform a perception task and can communicate and share information with other agents on the same task. 
Specifically, we propose a communication framework by learning both to construct communication groups and decide when to communicate. 
We demonstrate the generalizability of our framework on two different perception tasks and show that it significantly reduces communication bandwidth while maintaining superior performance. 

\end{abstract}

%%%%%%%%% BODY TEXT
%%%%%%%%% BODY TEXT
\section{Introduction}

Remarkable progress has been achieved for single-agent perception and recognition, where one or more sensor modalities are used to perform object detection~\cite{redmon2016you,Redmon_2017_CVPR,Lin_2017_ICCV} and segmentation~\cite{chen2018encoder,He_2017_ICCV,kirillov2019panoptic}, depth estimation~\cite{godard2017unsupervised,Zhou_2017_CVPR,godard2019digging}, and various other scene understanding tasks. 
However, in many applications, such as robotics, there may be multiple agents distributed in the environment, each of which has local sensors. 
Such multi-agent systems are advantageous in many cases, for example, to increase coverage across the environment or to improve robustness to failures. 
\begin{figure}[t]
    \begin{center}
    \centerline{\includegraphics[width=\linewidth]{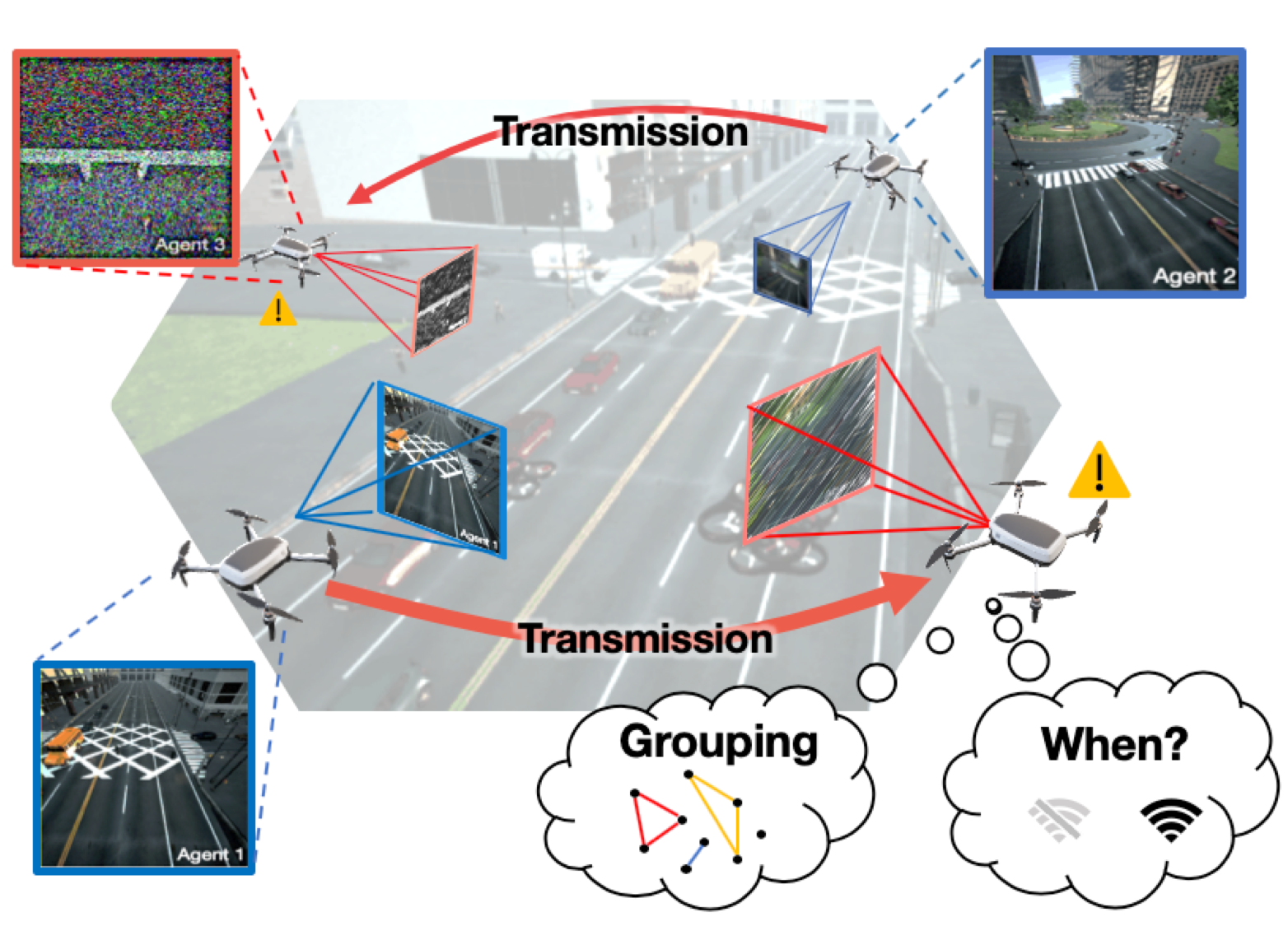}}
    \caption{
        \textbf{Illustration of multi-agent collaborative perception.} We construct a multi-agent perception system to improve the agent-wise perception accuracy and reduce the transmission bandwidth. Each agent learns to construct \textit{communication groups} and decide \textit{when to communication}.}
    \end{center}
    \vspace{-10mm}
\end{figure}

Thus, we tackle the problem of \textit{multi-agent collaborative perception}, an under-studied topic in the literature, where multiple agents are able to exchange information to improve overall accuracy towards perception tasks (\textit{e.g.}, semantic segmentation or object recognition).
One major challenge for multi-agent collaborative perception is the transmission bandwidth, as high bandwidth results in network congestion and latency in the agent network.
We therefore investigate the scenario where information across all agents (and hence sensors) is not available in a centralized manner, and agents can only communicate through bandwidth-limited channels. 
We also consider several challenging scenarios where some sensor data may be uninformative or degraded. 

Prior works on learning to communicate~\cite{sukhbaatar2016learning,foerster2016learning} mainly address decision-making tasks (rather than for improving perception) under simple perceptual environments. 
In addition, these methods also do not consider bandwidth limitations: They learn to communicate across a fully-connected graph (i.e. all agents communicate with each other through broadcasts).
Such methods cannot scale as the number of agents increases. 
Similarly, since all information is broadcast there is no decision of \textit{when} to communicate conditioned on the need. 
An agent does not need to consume bandwidth when the local observation is sufficient for the prediction. 
When messages sent by other agents are degraded or irrelevant, communication thus could be detrimental to the perception task.   

In this paper, we propose a learning-based communication model for collaborative perception.
We specifically view the problem as learning to construct the communication group (i.e. each agent decides what to transmit and which agent(s) to communicate with) and to decide when to communicate without explicit supervision for such decisions during training. 
In contrast to broadcast-based methods (e.g. TarMac~\cite{das2019tarmac}) and inspired by the general attention mechanisms, our method decouples the stages of communication and this allows for \textit{asymmetric message and key sizes}, reducing the amount of transmitted data.

Our method can be generalized to several downstream vision tasks, including multi-agent collaborative semantic segmentation (dense prediction) and multi-agent 3D shape recognition (global prediction). 
Our model is able to be trained in an end-to-end manner with only supervision from downstream tasks (e.g. ground-truth masks for segmentation and class labels for image recognition) and without the need for explicit ground-truth communication labels. 

We demonstrate across different tasks that our method can perform favorably against previous works on learning to communicate while using less bandwidth. 
We provide extensive analyses, including trade-offs between message and query sizes, the correlation between ground-truth key and predicted message, and visualization of the learned communication groups.

Our contributions are listed as follows:
\begin{itemize}[topsep=0pt,itemsep=-1ex,partopsep=1ex,parsep=1ex,labelindent=0.0em,labelsep=0.2cm,leftmargin=*]
\item We address the under-explored area of collaborative perception, which is at the intersection of perception, multi-agent systems, and communication.
\item We propose a unified framework that learns both how to construct communication groups and when to communicate. It does not require ground truth communication labels during training, and it can dynamically reduce bandwidth during inference.
\item Our model can be generalized to several down-stream tasks, and we show through rigorous experimentation that it can perform better when compared with previous works investigating learned communication.
\item  We provide a collaborative multi-agent semantic segmentation dataset, AirSim-MAP, where each agent has its own depth, pose, RBG images, and semantic segmentation masks. This dataset allows researchers to further investigate solutions to multi-agent perception. 
\end{itemize}

\section{Related works}
\begin{figure}[t]
    \vspace{1mm}
    \begin{center}
    \centerline{\includegraphics[width=\linewidth]{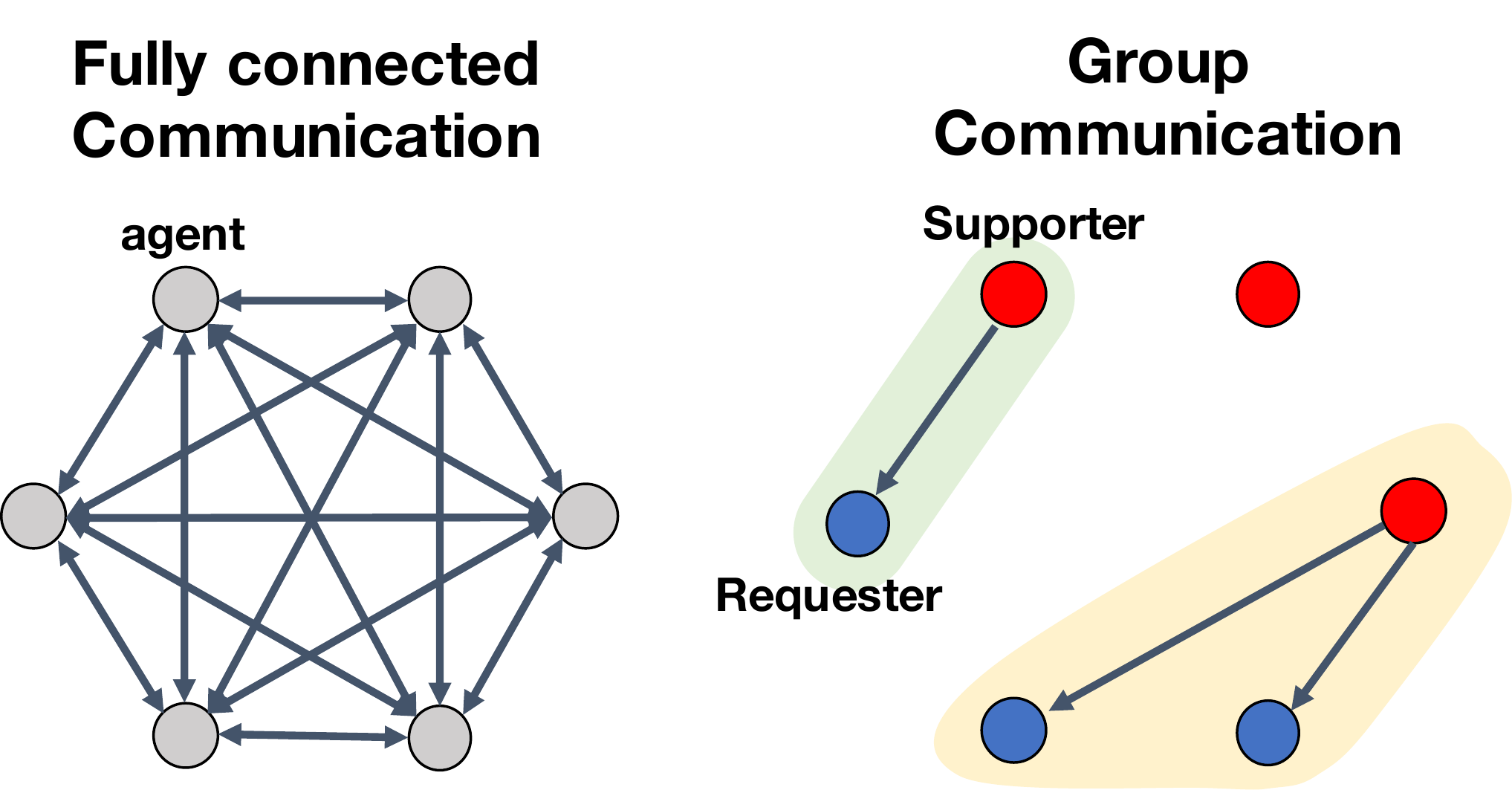}}
    \caption{
        \textbf{Fully connected versus group communication.} Fully connected communication results in a large amount of bandwidth usage, growing on the order of $\mathcal{O}(N^2)$, where $N$ represents the number of agents in a network. Group communication is able to prune irrelevant connections and can substantially reduce the overall network complexity.
        }
        \label{fig:com_diff}
    \end{center}
    \vspace{-6mm}
\end{figure}

\textbf{Learning to communicate.}
Communication is an essential component for effective collaboration, especially for multi-agent decision-making and perception tasks.  
Early works~\cite{tan1993multi,melo2011querypomdp} facilitated information flow and collaboration through pre-defined communication protocols. Similarly, auction-based methods~\cite{li2010auction,qureshi2008smart} for camera grouping use several assumptions (e.g., static cameras) and heuristic rules to decide the agents’ communication. 
However, such rigid protocols do not evolve with dynamic environments and do not easily generalize to complex environments.  
Thus, in recent years, several multi-agent reinforcement learning (MARL) works have explored learn-able interactions between agents.  
For example, assuming full cooperation across agents, each agent in CommNet~\cite{sukhbaatar2016learning} broadcasts its hidden state to a shared communication channel so that other agents can decide their actions based on this integrated information.  
A similar scheme was proposed by Foerster~\textit{et al.}~\cite{foerster2016learning}, where agents instead communicate via learned, discrete signals.  
To further leverage the interactions between agents, Battaglia~\textit{et al.}~\cite{battaglia2016interaction} and Hoshen~\cite{hoshen2017vain} integrate kernel functions into CommNet.  
Additionally, several works have addressed communication through recurrent neural networks (RNN). 
For example, DIAL~\cite{foerster2016dial} uses an RNN to derive the individual Q-value of each agent based on its observation and the messages from other agents.
BiCNet~\cite{peng2017multiagent} connects all agents with a Bi-directional LSTM to integrate agent-specific information, and ATOC~\cite{jiang2018learning} additionally applies an attention unit to determine what to broadcast to the shared channel.  
Although substantial progress has been made by several MARL works, most experimental tasks are built on simplistic 2D-grid environments where each agent observes low-dimensional 2D images. 
As mentioned in Jain~\textit{et al.}~\cite{jain2019two}, studying agents' interactions in simplistic environments does not permit study of the interplay between perception and communication. 

\begin{figure*}[t]
    \vspace{1mm}
    \begin{center}
        \centering
   \begin{picture}(600,160)
     \put(15,10){\includegraphics[width=\linewidth]{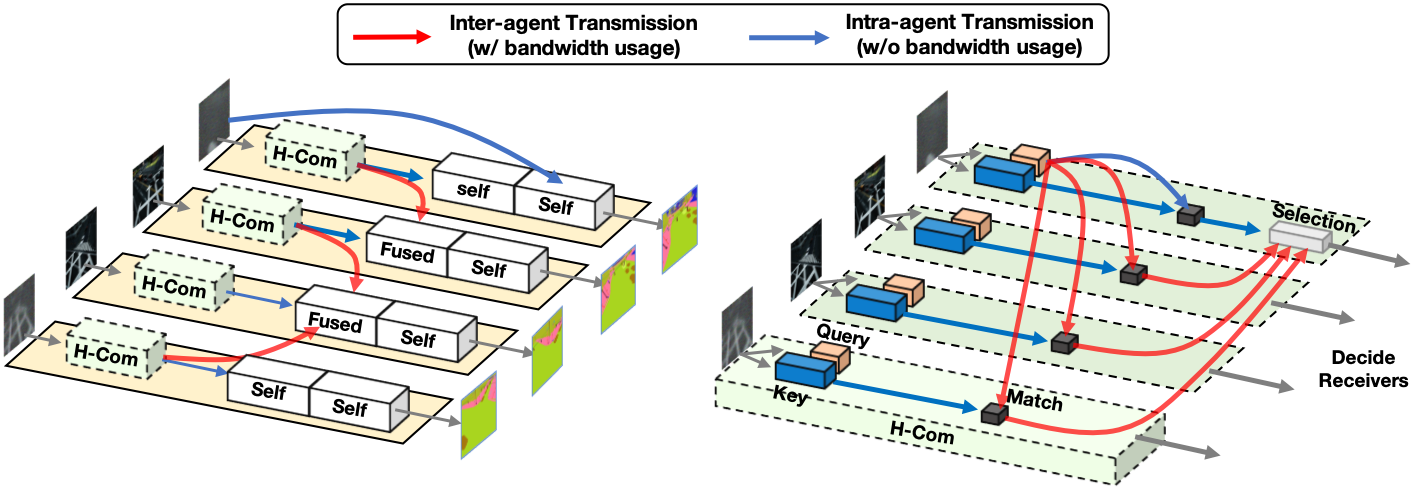}}
     \put(127,0){(a)}
     \put(383,0){(b)}
   \end{picture}
        \caption{
        Our (a) \textbf{Multiple-Request Multiple-Support Model} and its (b) \textbf{Handshake-Communication (H-Com) Module}.}
        \label{fig:overview_model}
    \end{center}
\end{figure*}

Recent works have proposed to construct communication groups based on pre-defined rules~\cite{jiang2018learning,jiang2018graph} or a unified communication network~\cite{singh2018learning,sukhbaatar2016learning,hoshen2017vain,peng2017multiagent,singh2018learning,das2019tarmac}. 
With these techniques, bandwidth usage during communication increases as the number of agents scales up. While Who2com~\cite{liu2020who2com} uses a handshake communication to reduce the bandwidth usage, this model assumes all agents \textit{always} need to communicate with one of the other agents. 
This results in the waste of bandwidth consumption and cannot prevent the issue of detrimental messages. 
In contrast, our proposed framework alleviates these problems by learning to decide when to communicate and to create communication groups.

\textbf{Attention mechanisms.} 
Attention mechanisms have been widely used in recent learning-based models.
In a nutshell, the attention mechanism can be viewed as a soft fusion scheme to weight different values based on a similarity between query and keys. 
A few noticeable and widely used attention mechanisms are \textit{additive}~\cite{bahdanau2014neural}, \textit{scale dot-product}~\cite{vaswani2017attention}, and \textit{general}~\cite{luong2015effective}.
One key finding of our work is that the \textit{general} mechanism allows for asymmetric queries and keys, which makes it especially suitable for tasks with bandwidth considerations: An agent's transmitted query message can be smaller than its retained key, and hence its overall bandwidth consumption can be reduced.

\section{Method}

The goal of our proposed model is to address the multi-agent collaborative perception problem, where an agent is able to improve its perception by receiving information from other agents. In this paper, we are specifically interested in \textit{learning to construct communication groups} and \textit{learning when to communicate} in a bandwidth-limited way.

\subsection{Problem Definition and Notation}
We assume an environment consisting $\bm{N}$ agents with their own observations $\bm{X}=\{\bm{x}_n\}_{n=1,...,\bm{N}}$.
Among those agents, some of them are degraded $\tilde{\bm{X}} = \{\tilde{\bm{x}}_l\}_{l=1,..,\bm{L}}$, and the set of degraded agents is a subset of all agents $\tilde{\bm{X}}\subset\bm{X}$. 
Each agent outputs the prediction of perception task $\tilde{\bm{Y}}=\{\bm{\tilde{y}}_n\}_{n=1,...,\bm{N}}$ with the proposed communication mechanism. 
Note that each agent is a \textbf{\textit{requester}} and \textbf{\textit{supporter}} simultaneously. 
However, which agents are degraded is \textbf{unknown} in our problem setting.

\subsection{Communication Groups Construction}

% motviation
As demonstrated in Figure ~\ref{fig:com_diff}, previous works on learning to communicate applied fully-connected communication for information exchange across agents. This framework results in a large amount of bandwidth usage and is difficult to scale up when the number of agents increases. 

% our method
To reduce the network complexity and bandwidth usage, inspired by communication network protocol~\cite{kurose2005computer}, we propose a two-step group construction procedure: we first apply the handshake communication~\cite{liu2020who2com} to determine the weights of connections, and we further prune the less important connections with an activation function.  

To start constructing a communication group, we apply a three-stage handshake communication mechanism~\cite{liu2020who2com}, which consists of three stages: request, match, and select. Agent $i$ first compresses its local observations $\bm{x}_i$ into a compact query vector $\bm{\mu}_i$ and a key vector $\bm{\kappa}_i$:
\begin{equation}\label{eq:key_query_generator}
     \bm{\mu}_i = G_q(\bm{x}_i; \theta_q)\in\mathbb{R}^Q, \quad \bm{\kappa}_i = G_k(\bm{x}_i;\theta_k)\in\mathbb{R}^K,
\end{equation}
where $G_q$ is a query generator parameterized by $\theta_q$ and $G_k$ is a key generator parameterized by $\theta_k$. We further broadcast the query to all of other agents, and note that this only causes limited amount of bandwidth transmission as all queries are compact compared to the high-resolution images. To decide who to communicate with, we compute a matching score $\bm{m}_{i,j}$ between an agent $i$ (as a requester) and an agent $j$ (as a supporter),
\begin{equation}\label{eq:matching}
    \bm{m}_{i,j} = \Phi(\bm{\mu}_i, \bm{\kappa}_j), \quad \forall i\neq j,
\end{equation}
where $\Phi(\cdot,\cdot)$ is a learned similarity function which measures the correlation between two vectors. The matching score $\bm{m}_{i,j}$ implies correlation between agent $i$ and $j$, and intuitively this value also represents the amount of information the supporting agent $j$ can provide for the requesting agent $i$. 

However, the above method does not learn ``when" to communicate, and it results in wasted bandwidth when an agent has sufficient information and the communication is not necessary. An ideal communication mechanism is to switch on transmission when the agent requires information from other agents to improve its perception ability, while it should also switch off the transmission when it has sufficient information for its own perception task. Toward this end, inspired by self-attention mechanism~\cite{cheng2016long}, we use the correlation between the key and query from the same agent to determine if the agent potentially requires more information and thus learn when to communicate,
% equtation for self-attention 
\begin{equation}\label{eq:matching}
    \bm{m}_{i,i} = \Phi(\bm{\mu}_i, \bm{\kappa}_i).
\end{equation}
Note that $\hat{\bm{m}}_{i,i}\approx1$ represents that the agent has sufficient information and does not need communication for perception tasks.

In order to minimize bandwidth usage during transmission, we further propose an asymmetric message method, which compresses the query into an extremely low-dimensional vector (which is transmitted) while keeping a larger size for the key vector (which is not transmitted). Once extremely compact queries are passed to receiver agents, we use a scaled general attention~\cite{luong2015effective,vaswani2017attention} to compute the correlation between agent $i$ and agent $j$:
\begin{equation}\label{eq:matching_detail}
   \Phi(\bm{\mu}_i, \bm{\kappa}_j)= \frac{\bm{\mu}_i^T \bm{W}_g \bm{\kappa}_j}{\sqrt{K}},
\end{equation}
where $W_g \in \mathbb{R}^{Q \times K}$ is a learnable parameter to match the size of query and key, and $Q$ and $K$ are dimension of query and key respectively.

% MIMO
Based on the above self-attention and cross-attention mechanism across all queries and keys, we thus derive the matching matrix $\bm{M}$:
\begin{equation}\label{eq:matching}
\bm{M} = 
 \bm{\sigma}(\begin{pmatrix}
  \bm{m}_{1,1} & \bm{m}_{1,2} & \cdots & \bm{m}_{1,N} \\
  \bm{m}_{2,1} & \bm{m}_{2,2} & \cdots & \bm{m}_{2,N} \\
  \vdots  & \vdots  & \ddots & \vdots  \\
  \bm{m}_{N,1} & \bm{m}_{N,2} & \cdots & \bm{m}_{N,N} 
 \end{pmatrix})\in\mathbb{R}^{\bm{N}\times \bm{N}},
\end{equation}
where $\bm{\sigma}(\cdot)$ is a row-wise softmax function.

To construct the communication groups, we prune the less connections with an activation function:
\begin{equation}\label{eq:matching}
\bar{\bm{M}} = \bm{\Gamma}(
   \bm{M} ;\delta),
\end{equation}
where $\bm{\Gamma}(\cdot;\delta)$ is an element-wise function, which zeros out the elements smaller than $\delta$. (We set $\delta=\frac{1}{N}$ in our experiment.)

\begin{figure}[t]
    \vspace{1mm}
    \begin{center}
    \centerline{\includegraphics[width=0.9\linewidth]{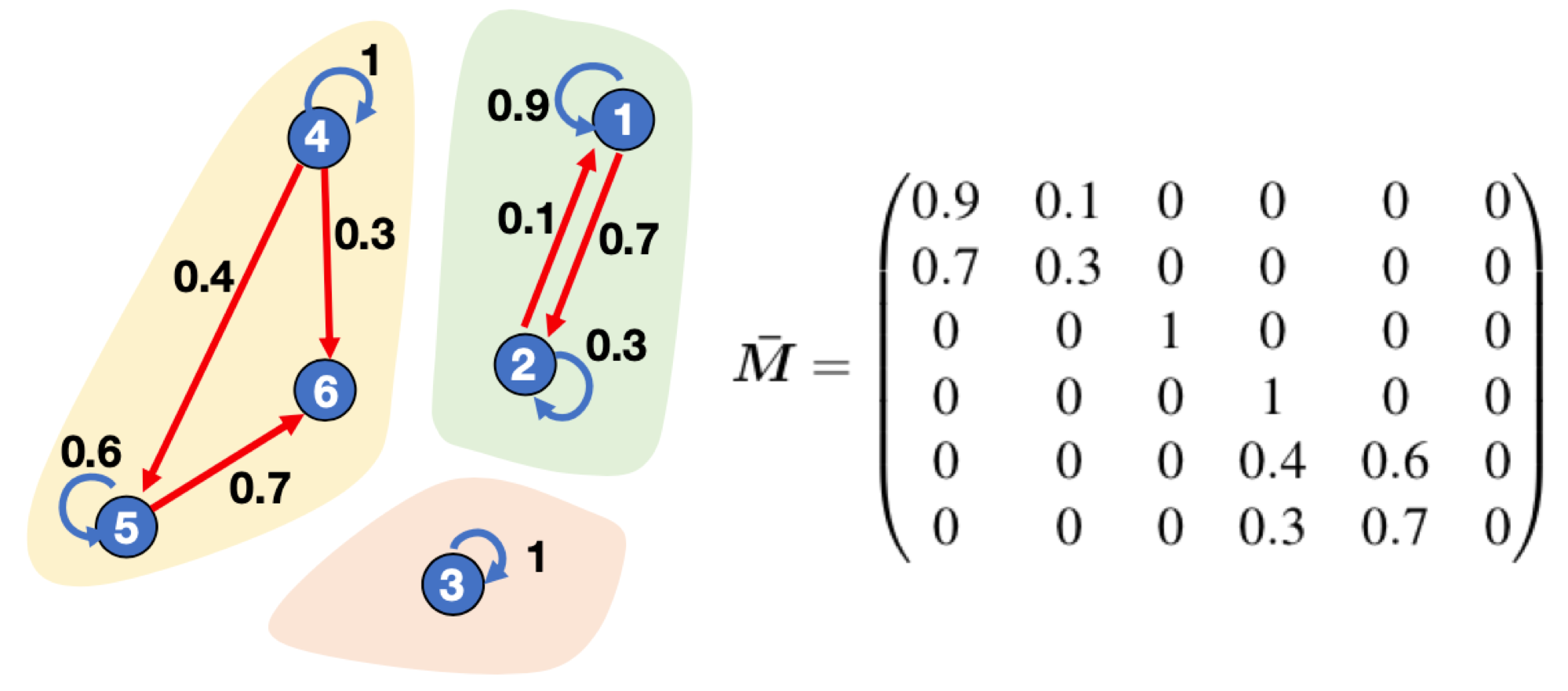}}
    \caption{
        \textbf{An example of our constructed communication groups and the corresponding adjacency matrix.} Blue arrow indicates the intra-agent transmission without bandwidth consumption, and red arrow represents the inter-agent transmission with bandwidth consumption.}
        \label{fig:graph_mat}

    \end{center}
    \vspace{-6mm}
\end{figure}

The derived matrix $\bar{\bm{M}}$ can be regarded as the adjacency matrix of a directed graph, where the entries of the matrix indicate when to communicate and non-entries represent who to communicate with as demonstrated in Figure~\ref{fig:graph_mat}. 
Each row of the matrix represents how a receiver agent collects information from different supporting agents, and each column of the matrix is how one supporter sends its own information to different requesting agents.  

As shown in Figure~\ref{fig:overview_model}, once a requesting agent collects the information from its linked supporting agents, the requesting agent $i$ integrates its local observation and the compressed visual feature maps from supporters based on the matching scores:
\begin{align}\label{eq:softmax}
        \bm{\hat{y}}_i=D([{\bm{f}}_i; \bm{f}_{i}^{inf}];\theta_d), \quad
        \bm{f}_{i}^{int} = \sum_{\substack{j=1 \\  \bm{\bar{m}}_{i,j}\neq 0 }}^{\bm{N}}\bm{\bar{m}}_{i,j}\bm{f}_j,
\end{align}
where $\bm{D}$ is an perception task decoder parameterized by $\theta_d$, $\bm{\bar{m}}_{i,j}$ is the element located in
$i$-th row and $j$-th of the matrix $\bm{\bar{M}}$, $\bm{f}_i=\bm{E}(x_i;\theta_e)$ is an feature map of agent $i$ encoded by an image encoder $\bm{E}$, $[\cdot;\cdot]$ is concatenation operation along channel dimension. It is worth noting that our perception task decoder is not limited for specific vision tasks, and we demonstrate our communication framework can be generalized to different visual tasks in our experiments. 

\subsection{Learning of Communication}
Our learning strategy follows the centralized training and decentralized inference procedure~\cite{lowe2017multi}. 
Precisely, all agents are able to access all local observations of agents in the training stage, while each agent can only observe its own local observation in the inference stage. Our goal is to learn a bandwidth-efficient communication mechanism, so that in the inference stage, our proposed model is able to perform multi-agent collaborative perceptions in a bandwidth-limited and distributed manner. 

We follow the aforementioned handshake communication to compute the matching matrix $M$, and we weight the agents' feature maps based on the matching matrix $M$ and further integrate them as: 
\begin{align}\label{eq:training_softmax}
        \bm{f}_{i}^{all} = \sum_{j=1}^{\bm{N}}\bm{\tilde{m}}_{i,j}\bm{f}_j,
\end{align}
where $\bm{\tilde{m}}$ the element located in
$i$-th row and $j$-th of the matrix $\bm{M}$. 
Note that in the above equation $\bm{m}_{i,j}\bm{f}_j$ represents who to communicate with, and $\bm{m}_{i,i}\bm{f}_i$ indicates when to communicate.
Then, a client agent $i$ combines its own feature map $f_i$ and the integrated feature $\bm{f}_{i}^{all}$ to compute the prediction for downstream visual tasks,
\begin{align}\label{eq:training_softmax}
        \bm{\tilde{y}}_i=D([{\bm{f}}_i; \bm{f}_{i}^{all}];\theta_d), \quad
\end{align}

In order to train our model, we use the label for downstream tasks (\textit{e.g.,} segmentation masks) as supervision, we compute the loss as:
\begin{align}\label{eq:loss}
\mathcal{L} = \mathcal{H}(\bm{y}_j, \bm{\tilde{y}}_j),
\end{align}
where $\mathcal{H}(\cdot,\cdot)$ can be the objective function for any downstream visual tasks (\textit{e.g.} pixel-wise cross-entropy for segmentation tasks or cross-entropy for recognition tasks). We later update the weights of our model $\Theta=(\theta_k,\theta_q, \theta_e, \theta_d)$ using the above loss in an end-to-end manner.

\section{Experiment}
We evaluate the performance of our proposed framework on two distinct perception tasks: collaborative semantic segmentation and multi-view 3D shape recognition.  
% The first task is divided into three cases.  All experimental cases are summarized in Figure ~\ref{fig:experimental_cases}.
% We will first introduce the datasets for both tasks and then briefly describe the baseline methods.  

\subsection{Experimental Cases and Datasets}
% \begin{table*}
% \caption{\textbf{Comparison of semantic segmentation datasets\yencheng{need to polish this table}}}
% \label{tab:dataset}
% \centering
% \begin{tabular}{cccccc}
% \toprule
% & $\#$ of Cams & $\#$ of Agents & Depth & RGB & Pose info \\
% \midrule
% KITTI~\cite{geiger2013vision} & 2 & 1 & single & single & ? \\
% CityScapes~\cite{cordts2016cityscapes} & 2 & 1 & single & single & ? \\
% Oxford~\cite{RobotCarDatasetIJRR} & 4 & 1 & single & single & ?\\
% Drive360~\cite{hecker2018end} & 8 & 1 & single & single & ? \\
% AirSim-MAP & 6 & 6 & multiple & multiple & multiple\\
% \bottomrule
% \end{tabular}
% \end{table*}

% 
\subsubsection{Collaborative Semantic Segmentation}

Our first task is collaborative 2D semantic segmentation of a 3D scene. Given observations (an RGB image, aligned dense depth map, and pose) from several mobile agents, the objective of this task is to produce an accurate 2D semantic segmentation mask for each agent.

Since current semantic segmentation datasets~\cite{geiger2013vision,cordts2016cityscapes,RobotCarDatasetIJRR,hecker2018end} only provide RGB images and labels captured from the perspective of single agent, we thus use AirSim simulator~\cite{airsim2017fsr} to collect our AirSim-MAP (Multi-Agent Perception) dataset.
For this dataset, we fly a swarm of five to six drones with different yaw rates through a series of waypoints in the AirSim ``CityEnviron'' environment. 
We record pose, depth maps, and RGB images for each agent. Note that we also provide semantic segmentation masks for \textbf{all} drones.  

We consider three experimental cases within this task.  We refer to the agent attempting segmentation as the \textbf{requesting} agent, and all other agents as the \textbf{supporting} agents.  Details for each case are listed as follows: 

\noindent
\textbf{Single-Request Multiple-Support (SRMS)} 
This first case examines the effectiveness of communication for a single requesting agent under the assumption that if an agent is degraded, then its original, non-degraded information will be present in one of the supporting agents.  We include a total of five agents, of which only one is selected for possible degradation.  We add noise to a random selection of $50\%$ of this agent's frames, and we randomly replace one of the remaining agents with the \textit{non-degraded} frame of the original agent.  Note that only the segmentation mask of the original agent is used as supervision.  

\noindent
\textbf{Multiple-Request Multiple-Support (MRMS)}
The second case considers a more challenging problem, where multiple agents can suffer degradation.  Instead of requiring a single segmentation output, this case requires segmentation outputs for all agents, degraded and non-degraded.  We follow the setup of the previous case, and we ensure that each of the several degraded requesting agents has a corresponding non-degraded image among its supporting agents.  

\noindent
\textbf{Multiple-Request Multiple-Partial-Support (MRMPS)}
The third case removes the assumption that there exists a clean version of the degraded view among the supporting agents.  Instead, the degraded agent must select the most informative view(s) from the other agents, and these views might have a variable degree of relevance.  Specifically, as the drone group moves through the environment, the images from each drone periodically and partially overlap with those of other drones.  Intuitively, the segmentation output of the requesting drone is only aided from the supporting drones that have overlapping views. 

\begin{table*}
\caption{\textbf{Experimental results on Multiple-Request Multiple-Support and Multiple-Request Multiple-Partial-Support.} Note that we evaluate these models with the metric of mean intersection of union (mIoU) and use MBytes per frame (Mbpf) and the averaged number of links per agent for measuring bandwidth.}
\label{tab:mimo_case1}
\centering
\resizebox{\linewidth}{!}{%
\begin{tabular}{ccccccccccc}
\toprule
& &  \multicolumn{4}{c}{Multiple-Request Multiple-Support} & & \multicolumn{4}{c}{Multiple-Request Multiple-Partial-Support} \\
 \cmidrule(lr){3-6}  \cmidrule(lr){8-11} 
& Models & Bandwidth (Mbpf / $\#$ of links) & Noisy & Normal & Avg.&  & Bandwidth  (Mbpf / $\#$ of links)& Noisy & Normal & Avg.  \\
\midrule
 & AllNorm & - & 57.85 & 57.74 & 57.80&  & - & 47.9 & 48.37 & 48.14  \\
\midrule

\multirow{5}{*}{ Fully-Connect. } & CatAll & 2.5 / 5 & 29.07 & 51.83 & 40.45 & & 2.0 / 4 & 26.86 & 45.27 & 36.07 \\
& AuxAttend & 2.5 / 5 & 33.69 & 56.27 & 44.98  & & 2.0 / 4 & 26.97 & 51.03 & 39.00\\
& CommNet~\cite{sukhbaatar2016learning} & 2.5 / 5 & 23.68 & 52.67 & 38.18 & & 2.0 / 4 & 26.56 & 49.07 & 37.82 \\
& TarMac~\cite{das2019tarmac} & 2.5 / 5 & 51.09 & 56.74 & 53.92 & & 2.0 / 4 & 29.78 & \textbf{51.39} & 40.59 \\
\midrule
\multirow{3}{*}{ Distri. } & RandCom & 0.5 / 1 & 21.22 & 52.74 & 36.98 & & 0.5 / 1 & 24.13 & 45.19 & 34.66 \\
& Who2com~\cite{liu2020who2com} & 0.5 / 1 & 31.96 & 56.11 & 44.04 &  & 0.5 / 1 & 26.97 & 50.71 & 38.84 \\
& Ours & \textbf{0.385 / 0.77} & \textbf{56.52} & \textbf{58.04} & \textbf{57.28} & & \textbf{0.55} / 1.08 & \textbf{30.38} & 51.26 & \textbf{40.82} \\
\midrule
 & OccDeg & - & 30.06 & 56.31 & 43.19&  & - & 25.2 & 46.74 & 35.97 \\
\bottomrule
\end{tabular}}
\end{table*}

\subsubsection{Multi-Agent 3D Shape Classification}
\begin{figure*}[t]
    \vspace{1mm}
    \begin{center}
    \centerline{\includegraphics[width=0.7\linewidth]{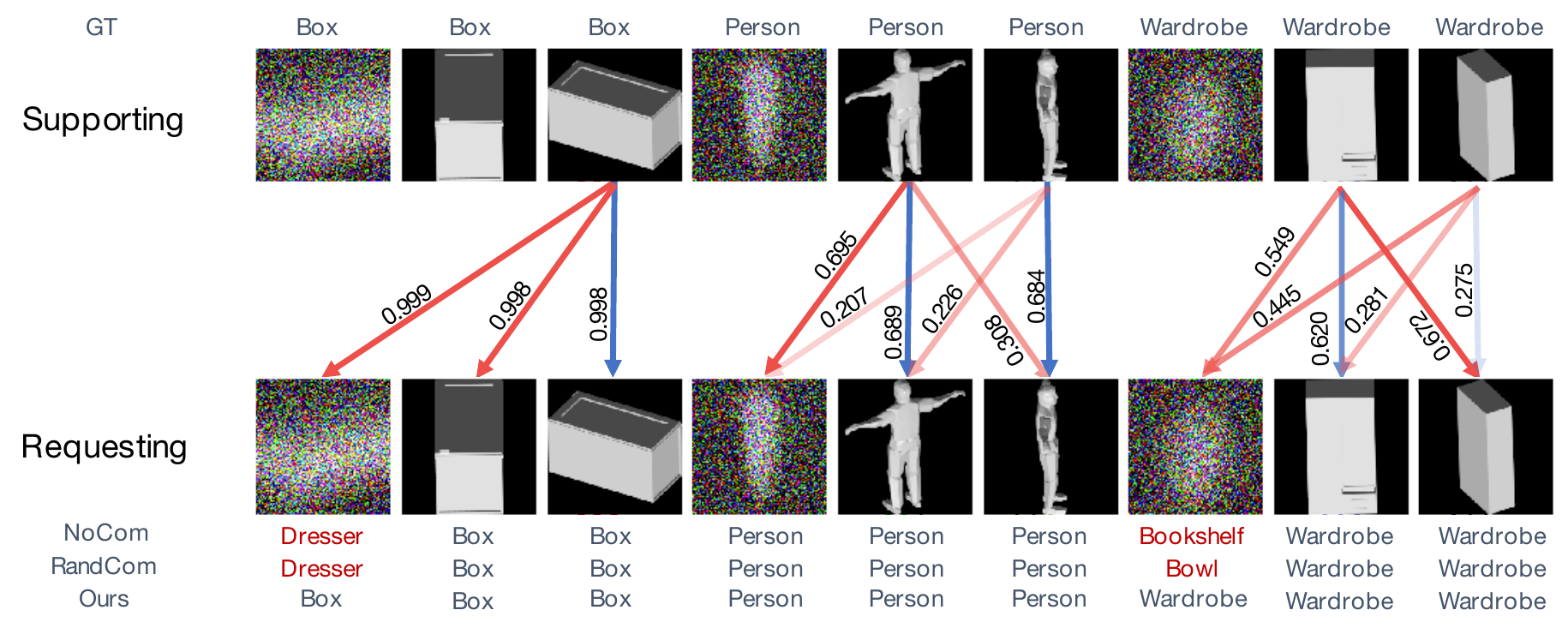}}
    \caption{
        \textbf{Bipartite communication graph between supporting and requesting agents.}  During the query phase, each requesting agent sends a low-dimensional query vector to all other agents (including itself) to establish communication.  Then during the transmission phase, supporting agents transmit their high-dimensional feature representations.  We visualize the flow of data during the transmission phase, where blue and red arrows refer to internal and external communication, respectively.  More prominent colors and larger numerical values indicate stronger feature weightings, whereas missing arrows represent the pruned links in the communication graph.  Note that these images are ordered for visualization purposes; the actual dataset is unordered, and each agent observes a random class with a random chance of degradation.}
        \label{fig:vis_graph}
    \end{center}
    \vspace{-8mm}
\end{figure*}

In addition to the semantic segmentation task, we also consider a multi-agent 3D shape classification task.  
For this experimental case, we construct a multi-agent variant of the \textbf{ModelNet 40} dataset~\cite{wu20153d}. 
The original dataset contains 40 common object categories from ModelNet with 100 unique CAD models per category and 12 different views of each model.  
However, our variant adds a communication group structure to the original dataset.  Specifically, we sample three sets of class-based image triplets.  
Each triplet corresponds to a randomly selected 3D object model and each triplet contains three randomly selected 2D views of its corresponding object model.  
To make this problem setting more challenging, we further degrade one image from each triplet. 
The objective of this task is to predict the corresponding object class for each agent by leveraging the information from all agents. 
Figure~\ref{fig:vis_graph} shows an example of the dataset in one trial with 9 agents.
This modified task is essentially a distributed version of the multi-view classification task~\cite{wu20153d}.

\begin{figure}[t]
    \vspace{1mm}
    \begin{center}
    \centerline{\includegraphics[width=\linewidth]{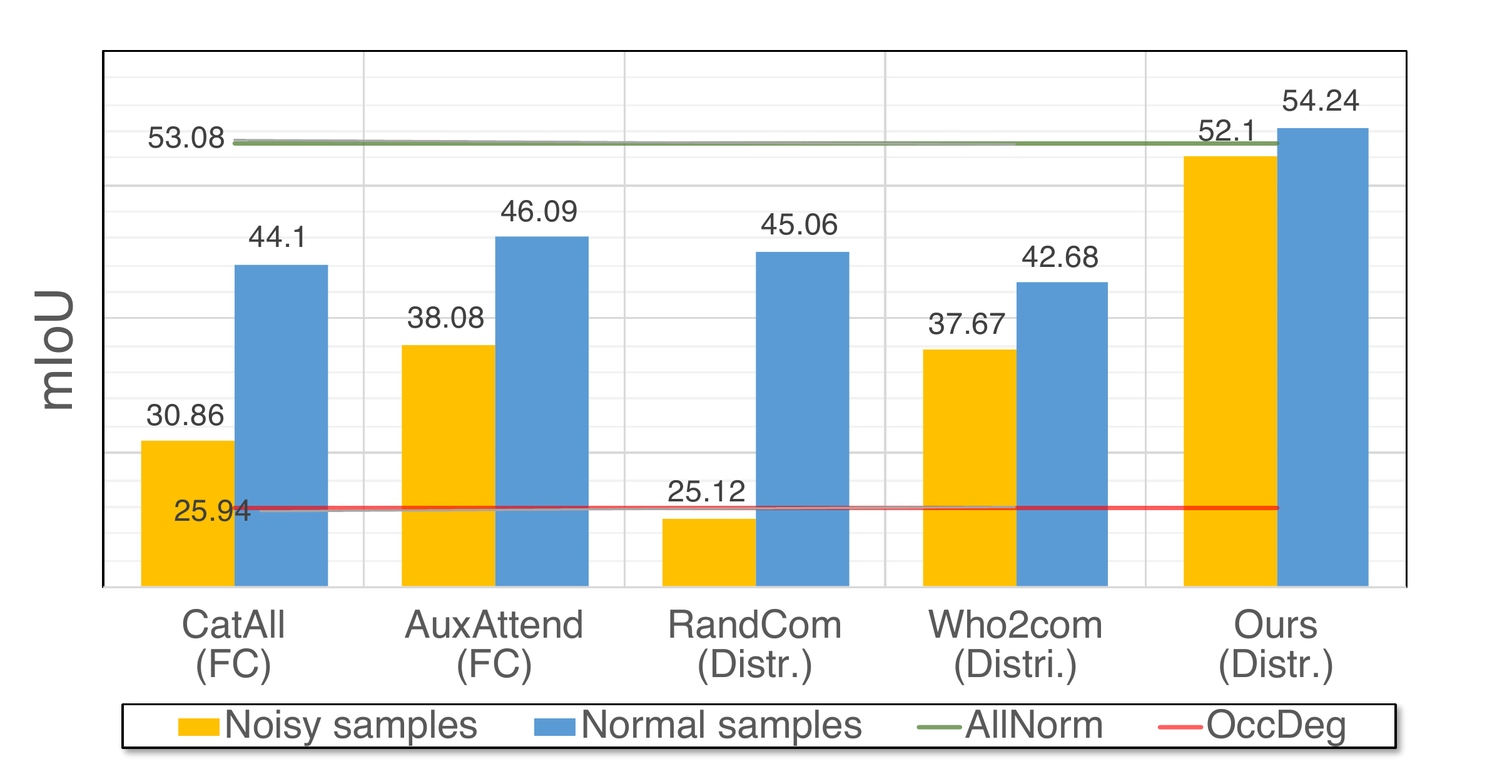}}
    \caption{
        \textbf{Experimental results of Single-Request Multiple-Support.}} 
        \label{fig:when2com_exp}
    \end{center}
    \vspace{-8mm}
\end{figure}

\subsection{Baselines and Evaluation Metrics}
Here we consider several fully-connected (FC) and distributed communication (DistCom) models as our baselines. FC models fuse all of the agents' observations (either weighted or unweighted) whereas DistCom models only fuse a subset of those observations.

\begin{itemize}[topsep=0pt,itemsep=-1ex,partopsep=1ex,parsep=1ex,labelindent=0.0em,labelsep=0.2cm,leftmargin=*]

\item \textit{CatAll} \textbf{(FC)} is a naive FC model baseline which concatenates the encoded image features of all agents prior to subsequent network stages.

\item \textit{Auxiliary-View Attention (AuxAttend;\textbf{FC})} uses an attention mechanism to weight auxiliary views from the supporting agents. 

\item  \textit{RandCom} \textbf{(DistCom)} is a naive distributed baseline which randomly selects one of other agents as a supporting agent. 

\item \textit{Who2com}~\cite{liu2020who2com} \textbf{(DistCom)} excludes self-attention mechanism such that it always communicates with one of the supporting agents. 

\item \textit{OccDeg and AllNorm} are baselines that employ no communication, i.e. each agent (view) independently computes the output for itself. For \textit{OccDeg} the data is degraded similarly as before, while in \textit{AllNorm} we use clean images for all views. These two serve as an upper and lower reference for comparison. 
\end{itemize}

We also consider communication modules of \textit{CommNet}~\cite{sukhbaatar2016learning}, \textit{VAIN}~\cite{hoshen2017vain}, and \textit{TarMac}~\cite{das2019tarmac} as our baseline methods for all multiple-outputs tasks. 
For a fair comparison, we use ResNet18~\cite{he2016deep} as the feature backbone for our and all mentioned  baseline models. For the 3D recognition task, we also add MVCNN~\cite{wu20153d} as a baseline.

We evaluate the performance of all the models with mean IoU on the segmentation task and prediction accuracy on the 3D shape recognition task. In addition, we report Bandwidth of all FC and DistCom models in Megabyte per frame (MBpf). To obtain MBpf, We add the size of the feature vectors which need to be transmitted to the requesters and size of keys broadcast to all supporters and multiply by the number of bytes required for storage.

\begin{table*}
\caption{\textbf{Experimental results on Multi-agent 3D Shape recognition.} We report the accuracy of the degraded split, and all methods perform similar results for the normal split ($\approx 83\%$). }
\label{tab:3d_shape}
\centering
\resizebox{\linewidth}{!}{%
\begin{tabular}{cccc|cccccc}
\toprule
 & OccDeg & AllNorm & RandCom & CatAll & MVCNN~\cite{wu20153d} & CommNet~\cite{sukhbaatar2016learning} & VAIN~\cite{hoshen2017vain} & TarMac~\cite{das2019tarmac} & Ours \\
\midrule
Degraded Split Accuracy ($\%$)&55.02 & 83.66 & 54.28 & 73.82 & 31.80 & 71.52 & 75.09 & 78.73 &\textbf{80.72}\\
Bandwidth (links/MBpf) &  - & - & 0.11 / 0.89 & 1 / 8 & 1 / 8 & 1 / 8 & 1 / 8 & 1 / 8 &  \textbf{0.176} / \textbf{1.32} \\
\bottomrule
\end{tabular}}
\end{table*}

\subsection{Quantitative results}

\noindent
\textbf{Single-Request Multiple-Support (SRMS)}
The goal of this case is to examine if our model is able to learn when to communicate and learn who to communicate with for a single requesting agent. 
Figure~\ref{fig:when2com_exp} shows the performance of our proposed model and several baseline models. 
Although most fully-connected methods can improve the prediction mIoU compared with \textit{NoCom}, they need to propagate all information in a fully-connected manner and thus require high bandwidth consumption. 
In contrast, our model reports higher prediction accuracy yet smaller bandwidth usage (Who2com~\cite{liu2020who2com}: $2$ MBpf; ours: $0.98$ MBpf).
Another observation is that our model is able to further improve compared with \textit{Who2com}~\cite{liu2020who2com}. 
This demonstrates the benefit of learning when to communicate, which reduces the waste of bandwidth and prevent detrimental message when the requesting agent has sufficient information and communication is not required. 

\noindent
\textbf{Multiple-Request Multiple-Support (MRMS)}
In this case, we further address a more challenging problem, where multiple agents suffer degradation. 
Each agent should (1) identify when it needs to communicate, (2) decide who to communicate with when it needs to, and (3) avoid the selection of noisy views from the supporting agents. 
We list the experiment results in the Table~\ref{tab:mimo_case1}\zk{Should move tables so that this is Table 1}. 
It can be seen that, when the requesting agents cannot prevent the selection of noisy supporting agents, both \textit{CatAll} and \textit{RandCom} perform even worse than \textit{NoCom}. 
This verifies our intuition that the information from the supporting agents is not always beneficial for the requesting agents, and selection of incorrect information may even hinder the prediction of the requesters.

With the use of attention mechanisms for weighting the feature maps from the supporting agents, both \textit{AuxAttend} and \textit{Who2com}~\cite{liu2020who2com} are able to prevent incorrect views from deteriorating performance and thus improve with respect to \textit{NoCom}, \textit{CatAll}, and \textit{RandCom}. 
However, without learning when to communicate, those models are forced to always request information from at least one supporting agent resulting in both poorer performance and unnecessary bandwidth usage. 

In addition to the above baseline methods, we also consider CommNet~\cite{sukhbaatar2016learning} and TarMac~\cite{das2019tarmac}. 
Even though CommNet integrates the information from other agents by using an average pooling mechanism, it does not improve the prediction of either degraded or non-degraded requesting agents because it indiscriminately incorporates all views. 

On the other hand, TarMac~\cite{das2019tarmac} is able to provide better results compared with the baseline models. 
However, TarMac uses one-way communication and results in large bandwidth usage which presents difficulty in the real scenario. 
On the contrary, our model is not only able to outperform it on both degraded and non-degraded samples, but also consumes less bandwidth by using our asymmetric query mechanism and pruning redundant connections within the network with the activation function.

\noindent
\textbf{Multiple-Request Multiple-Partial-Support (MRMPS)}
In this case, there is less chance to have completely overlapped observations between any two agents. 
This presents an inherent difficulty in the perception task because only incomplete information is available for the prediction. 
As shown in the right part of Table~\ref{tab:mimo_case1}, the performance improvement margin of all FC and DistCom models is smaller with respect to \textit{NoCom}, in comparison to more significant improvement observed in the previous scenario. 

Nonetheless, we observe that all methods exhibit a similar trend as the previous scenario. Our model is still able to maintain a similar prediction accuracy as fully-connected models, while we only use one-fourth bandwidth for communication across agents. This demonstrates the superior bandwidth-efficiency of our model.

\noindent
\textbf{Multi-agent 3D shape Recognition}
In order to demonstrate the generalization of our model, here we apply our model to the task of multi-agent 3D shape classification.
Table~\ref{tab:3d_shape} provides the quantitative evaluation on this task using our proposed model and other baselines, including \textit{VAIN}~\cite{hoshen2017vain}, \textit{CommNet}~\cite{sukhbaatar2016learning}, and \textit{TarMac}~\cite{das2019tarmac}. Our model is able to perform competitively compared with \textit{TarMac}~\cite{das2019tarmac} with only approximately one-eighth bandwidth usage. We also provide qualitative results in Figure~\ref{fig:vis_graph} to demonstrate the effectiveness of our model, which allows agents to communicate with the correct and informative agents.

\begin{figure}[t]
    \vspace{-6mm}
    \begin{center}
    \centerline{\includegraphics[width=70mm]{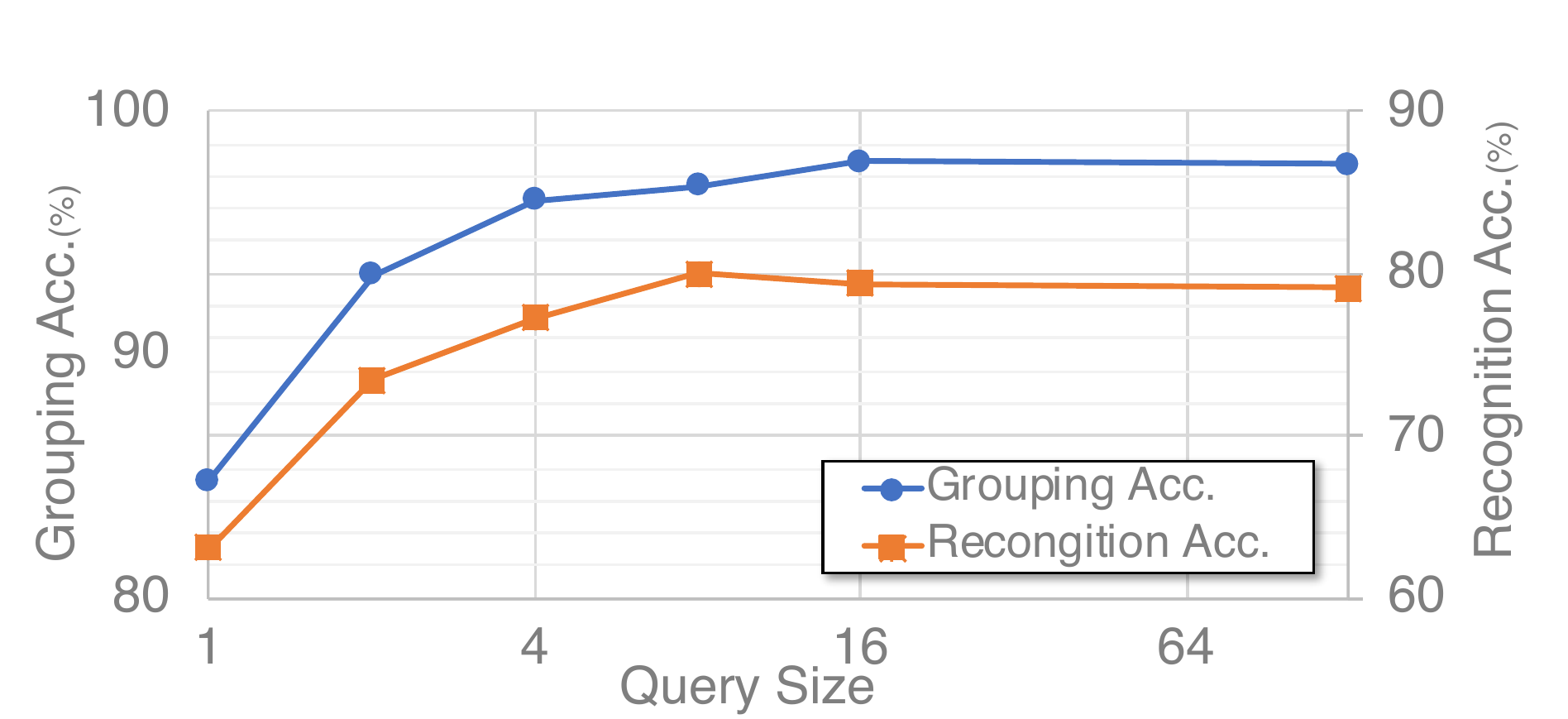}}
    \caption{
        \textbf{Ablation study on varying query size.}} 
        \label{fig:abl_msg}
    \end{center}
        \vspace{-10mm}
\end{figure}

\subsection{Analyses}

To investigate the source of our model's improvement over the baselines, we computed two selection accuracy metrics on the SRMS dataset, \textit{WhenToCom} and \textit{Grouping}. \textit{WhenToCom} accuracy measures how often communication between a requester and a supporter(s) is established and when it is needed; and \textit{Grouping} measures how often the correct group is created when there is indeed communication. We also comment on the trade-off between bandwidth and performance of communication by conducting a controlled experiment on the size of query and key on the 3D recognition dataset. 

\begin{figure}[t]
    \vspace{-6mm}
    \begin{center}
    \centerline{\includegraphics[width=0.8\linewidth]{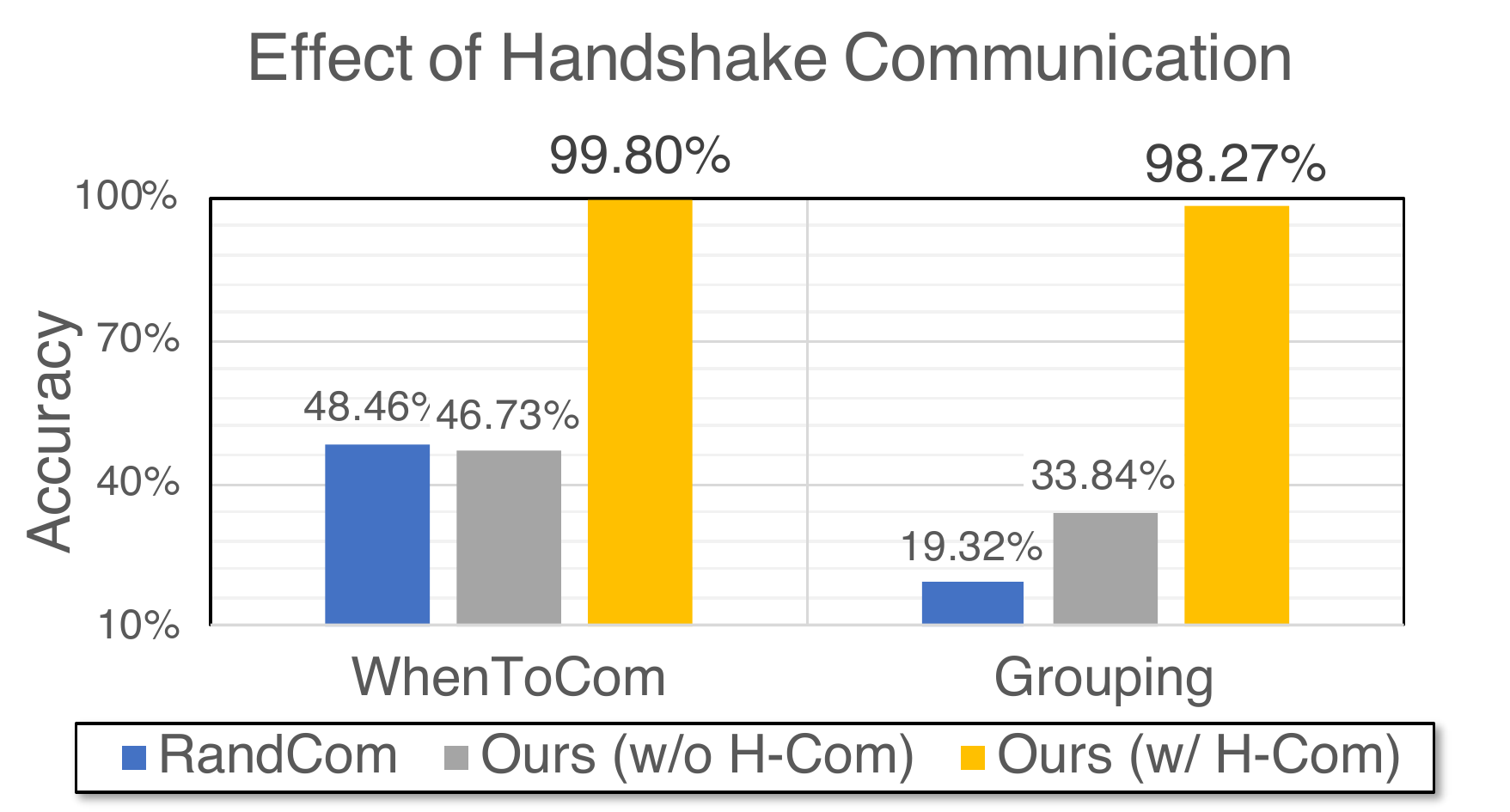}}
    \caption{
        \textbf{Effect of our proposed H-Com.} Handshake communication significantly improves the communication accuracy.} 
        \label{fig:hcom_effect}
    \vspace{-14mm}
    \end{center}
\end{figure}

\noindent
\textbf{Effect of handshake communication}
As demonstrated in Figure~\ref{fig:hcom_effect}, we conduct an ablation study on the proposed handshake communication. In the \textit{Ours (w/o H-Com)} model, we remove the handshake communication module, so that each agent only uses its local observations to compute both (1) the communication score and (2) its communication group.

We additionally provide the result of \textit{RandCom}.  We observed that our model with the proposed handshake communication offers a significant improvement over both \textit{RandCom} and our model without handshake communication. This finding demonstrates the necessity of communication for deciding when to communicate and who to communicate with. That is, an agent without communication cannot decide what information it needs and which supporter has the relevant information to help better perform on perception tasks.

Figure~\ref{fig:vis_graph} visualizes three examples from the 3D shape recognition task. Each agent clearly knows when communication is needed based on information provided by the supporters and its own observation. For example, in the first three box examples, the degraded agent on the left knows to select an informative view from the other agents; the non-degraded agent in the middle decides to select a more informative view even though it possesses sufficient information; and the third agent decides that communication is not needed because it has the most informative view among all. It is worth mentioning that all 9 views are provided to every agent and the agent needs to identify informative views and detrimental views based on the matching scores. 
\noindent
\textbf{Query and key size}
We further analyze the effect of query and key size on \textit{Grouping} accuracy and classification accuracy on the 3D shape classification task. 
We vary the query size from $1$ to $128$ with a fixed key size of $16$ as shown in Figure~\ref{fig:abl_msg}. We observe that both selection and classification accuracy improve as the message size increases. Our model can perform favorably with a message size of $4$. The same trend is also observed for key sizes. Most noticeably, we find that there exists asymmetry in query-key size. While the selection accuracy saturates at 16-dimensional query, selection accuracy consistently improves with increasing key size until 1024-dimensional key. Our model exploits this asymmetry to save bandwidth in communication while maintaining high performance.  

\section{Conclusion}
In this paper, we proposed a general bandwidth-efficient communication framework for collaborative perception. Our framework learns both how to construct communication groups and when to communicate. This framework can be generalized to several down-stream tasks including (but not limited to) multi-agent semantic segmentation and multi-agent 3D shape recognition. We demonstrated superior performance with lower bandwidth requirements across all compared methods.

%-------------------------------------------------------------------------
\section{Acknowledgement}
\label{sec:acknowledgement}
This work was supported by ONR grant N00014-18-1-2829.

{\small
\bibliographystyle{ieee_fullname}
\bibliography{egbib}
}

\end{document}